\documentclass[conference]{IEEEtran}
\IEEEoverridecommandlockouts
\usepackage{cite}
\usepackage{amsmath,amssymb,amsfonts}
\usepackage{algorithmic}
\usepackage{graphicx}
\usepackage{textcomp}
\usepackage{xcolor}
\def\BibTeX{{\rm B\kern-.05em{\sc i\kern-.025em b}\kern-.08em
    T\kern-.1667em\lower.7ex\hbox{E}\kern-.125emX}}
\begin{document}

\title{Supervised Manifold Learning via Random Forest Geometry-Preserving Proximities*\\

}

\author{\IEEEauthorblockN{Jake S. Rhodes}
\IEEEauthorblockA{\textit{Department of Statistics} \\
\textit{Brigham Young University}\\
Provo, Utah, USA \\
rhodes@stat.byu.edu}
}


\maketitle

\begin{abstract}
Manifold learning approaches seek the intrinsic, low-dimensional data structure within a high-dimensional space. Mainstream manifold learning algorithms, such as Isomap, UMAP, $t$-SNE, Diffusion Map, and Laplacian Eigenmaps do not use data labels and are thus considered unsupervised. Existing supervised extensions of these methods are limited to classification problems and fall short of uncovering meaningful embeddings due to their construction using order non-preserving, class-conditional distances. In this paper, we show the weaknesses of class-conditional manifold learning quantitatively and visually and propose an alternate choice of kernel for supervised dimensionality reduction using a data-geometry-preserving variant of random forest proximities as an initialization for manifold learning methods. We show that local structure preservation using these proximities is near universal across manifold learning approaches and global structure is properly maintained using diffusion-based algorithms.

\end{abstract}

\begin{IEEEkeywords}
supervised learning, manifold learning, random forest, data visualization, data geometry
\end{IEEEkeywords}

\section{Introduction}

Manifold learning algorithms are often used for exploratory data analysis. They are typically applied to noisy data in an attempt to find meaningful patterns or relationships across time, classes, or variables~\cite{Moon2019phate}. Most manifold learning approaches are unsupervised in that they do not use auxiliary information (e.g., data labels) in the embedding construction process. In many contexts, only unsupervised models are applicable as auxiliary information can be expensive or inaccessible. However, when available, label information can provide valuable insights into the data's intrinsic structure relative to the labels. Subjecting the embedding process to the use of auxiliary information can help to uncover a data geometry unattainable without labels. In this paper, we discuss weaknesses of current supervised manifold-learning approaches and show improvements on existing methods by applying a new variant of a random forest-based~\cite{Breiman2001randomforests} similarity measure in a manifold-learning setting. We use Geometry- and Accuracy-Preserving proximities (RF-GAP~\cite{rhodes2022geometry}) and demonstrate their ability to meaningfully encode a similarity measure and subsequent embedding that naturally incorporates labels. As opposed to distance or similarity measures which condition upon class labels to artificially exaggerate the separation of points of opposing classes, random forest proximities serve as a measure of similarity that uses labels (continuous, categorical, or otherwise) in a manner consistent with the model's learning. Additionally, forest-based proximities appropriately denoise the data, providing a meaningful metric or graph for the embedding process.


\section{Supervised Manifold Learning}\label{sec:supmanlean}

Manifold-learning algorithms use distance or similarity graphs to encode local data structure. For example, Isomap forms a $k$-nearest neighbor ($k$-NN) graph using Euclidean distance and seeks the shortest path between observations to approximate true geodesic distances upon which multi-dimensional scaling is applied~\cite{Tenenbaum2000isomap}. Diffusion Map (DM) uses a Gaussian kernel applied to a $k$-NN graph to form local similarities upon which eigendecomposition is applied~\cite{coifman2006dm}. T-distributed Stochastic Neighbor Embedding, or $t$-SNE, estimates probabilities as a normalized Gaussian kernel to define similarities between points. The Kullback-Leibler (KL) divergence between these and a lower-dimensional mapping is estimated via gradient descent to form the target embedding~\cite{vanDerMaaten2008tsne}. 

Each of these methods is unsupervised; data labels are not used in any part of the embedding process. However, supervised variants of these methods have been developed. Most of these supervised extensions of the algorithms adapt the existing algorithm at the distance- or similarity-learning level. In some cases, distances are rescaled~\cite{vlachos2002}, additively incremented~\cite{ridder2003slle}, or otherwise adapted conditionally upon class association~\cite{Ribeiro2008esiso}. Often, these dissimilarity measures can provide perfect linear separation where such discrimination is not possible using traditional classifiers. See Equation~\ref{eq:sisomap}, for an example of a class-conditional dissimilarity, where $D(.,.)$ denotes a distance function (e.g., Euclidean), $\beta$ is usually set as the average distance between points, $\alpha$ lessens separation between similar points of opposing classes, and $y_i$, $y_j$ are the respective labels of $x_i$ and $x_j$. This dissimilarity has been used to create supervised variants of $t$-SNE~\cite{hajderanj2019stsne}, Isomap~\cite{Ribeiro2008esiso}, Locally-Linear Embedding~\cite{ZHANG2009eslle}, and Laplacian Eigenmaps~\cite{jiang2009slapeig}. In each of these extensions, the within-class structure is partially maintained, but manifold structures are distorted at a global level as a result of exaggerated class separation. Such dissimilarity measures are order non-preserving bijections~\cite{hajderanj2021impactsupman}.

\begin{equation}
    \label{eq:sisomap}
    D'\left(\mathbf{x}_{i}, \mathbf{x}_{j}\right) = \left\{\begin{array}{ll} \sqrt{1-e^{\frac{-D^{2}\left(\mathbf{x}_{i}, \mathbf{x}_{j}\right)}{\beta}}} & y_{i}=y_{j} \\
    \sqrt{e^{\frac{D^{2}\left(\mathbf{x}_{i}, \mathbf{x}_{j}\right)}{\beta}}}-\alpha & y_{i} \neq y_{j}\end{array}\right.
\end{equation}

\begin{figure*}[!htb]
    \centering
    \includegraphics[width = .7\textwidth]{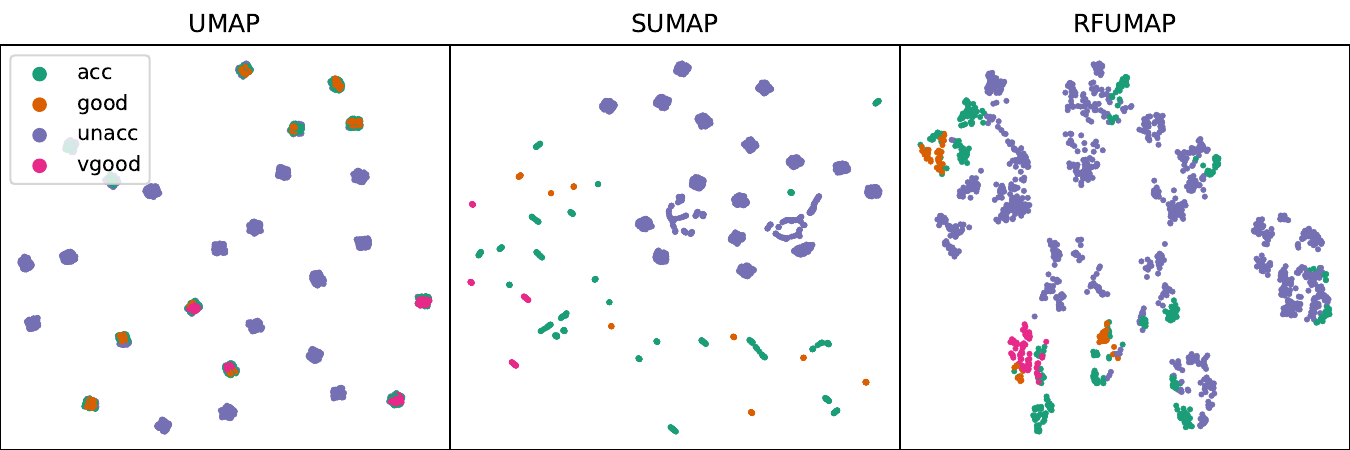}
    \caption{This figure provides UMAP-based 2-dimensional representations of the cars dataset~\cite{bohanec1988cars}. The left figure is the original UMAP~\cite{lel2018umap} implementation which shows distinct clusters of overlapping classes. A supervised version of UMAP~\cite{mcinnes2018umap-software} (center) provides perfect class separation as well as distinctive clusters, which does not coincide with a cross-validated $k$-NN accuracy of 0.899. The RF-GAP-based UMAP implementation (right) shows observational relationships which correspond to variable interactions within the dataset.}
    \label{fig:car-umap}
\end{figure*}

These approaches are problematic in several ways: (1) The class-conditional distances form an attempt to maintain with-class structure but cause disruption between classes. Artificial class separation diminishes inter-class relationships, thus distorting the global data structure. (2) The manifold disruption reduces the integrity of resulting downstream tasks. For example, classification tasks following dimensionality reduction can have unrealistically low error rates. (3) Class-conditional measures do not provide an avenue for continuously-valued labels. These extensions have not been adapted to regression problems. (4) These approaches are not extendable to new, unlabeled points (e.g., a test set used for subsequent predictions). To overcome each of these weaknesses, we propose the use of random forests~\cite{Breiman2001randomforests, rhodes2022geometry} to generate supervised similarities to be used in manifold learning.

\section{Random Forest Proximities}\label{sec:proximities}

Random forests~\cite{Breiman2001randomforests} provide a number of benefits for prediction problems that are supportive of metric learning. For example, random forests apply to both regression and classification problems, handle mixed variable types, provide an unbiased estimate of the generalization error, are insensitive to monotonic transformations, are relatively robust to outliers, and provide a natural way of assessing variable importance, ignoring noise variables in the presence of meaningful ones.

Random forests form an ensemble in binary-recursive decision trees each of which partitions a bootstrap sample of the training data. Observations within the bootstrap sample are called in-bag, while those in the training data not included in the sample are called out-of-bag, or OOB. Each partition forms a decision space used for classification or regression. The partitions naturally form a channel for generating similarity measures using data labels. These similarities are referred to in the literature as random forest proximities. 

Leo Breiman originally defined the proximity between two observations, $x_i$, and $x_j$, denoted by $p(x_i, x_j)$, as the number of terminal nodes they share across all trees, divided by the number of trees in the forest~\cite{BreimanRandomForest:Online}. This simple approach applies to all training points regardless of bootstrap status. As a result, proximities constructed on the training set tend to slightly overemphasize class segregation. To overcome this, an alternative formulation was derived to only calculates pair-wise similarities between points $x_i$ and $x_j$ using trees in which both of these observations are OOB~\cite{hastie2017elements}. Subsequently, these proximities combat overinflated class separation. However, it has been shown that OOB-only proximities do not fully benefit from the random forest's learning and are a noisier similarity measure~\cite{rhodes2022geometry}.

In~\cite{lin2006adaptiveNN}, the authors demonstrated that random forests behave like a nearest-neighbor regressor with an adaptive bandwidth. That is, random forest predictions (in the regression context) can be determined as a weighted sum using a kernel function, as shown in Equation~\ref{eq:weighted_sum}.

\begin{equation}\label{eq:weighted_sum}
    \hat{y}_i(k) = \sum_{j \ne i} k(x_i, x_j)y_j
\end{equation}

Here, $k$ is a weighted kernel function determined by the number of training examples sharing a terminal node with $x_i$. This is comparable to other kernel methods, such as the SVM, which uses a kernel to define similarity and ultimately the decision space. Ideally, random forest proximities should serve as a kernel capable of mimicking random forest predictions. Using normalized proximities as weights can serve as a test for the proximities' consistency with the forest's learning. In this regard, existing random forest proximity formulations do not adequately incorporate the forest's learning~\cite{rhodes2022geometry}.

Both the original formulation~\cite{BreimanRandomForest:Online}, as well as that using only OOB observations~\cite{hastie2017elements}, are not capable of reconstructing the random forest predictions for two reasons: (1) Random forests train on a set of bootstrap (in-bag) samples and predict on another set (the OOB samples or a test set). The original formulation doesn't discriminate between in-bag or OOB observations, and the OOB proximity definition does not use in-bag samples in their construction. (2) Decision tree voting takes into account the number of in-bag observations within a given terminal node, while the proximities are constructed without regard to the number of ``voting points"~\cite{lin2006adaptiveNN}. To construct proximities that serve as a kernel for random forest prediction, these two points must be accounted for.

In~\cite{rhodes2022geometry}, the authors propose a new proximity formulation capable of reconstructing random forest OOB and test predictions as a kernel method. They call these proximities Random Forest-Geometry- and Accuracy-Preserving proximities (RF-GAP) and show improvement across multiple applications using this new definition, including data imputation, outlier detection, and visualization via MDS. 

The RF-GAP proximities, however, do not form a proper kernel function as originally defined. They are asymmetric and self-similarity is defined to be 0 to account for the kernel prediction problem. To overcome this, we normalize the similarities to set the maximum similarity to 1, symmetrize the proximities, and define the diagonal entries to be 1. In doing so, the proximities can serve in any capacity which requires a kernel matrix. Using this modified RF-GAP formulation, these proximities serve as a similarity measure that overcomes the weaknesses of the class-conditional supervised distances in the following ways: (1) Rather than conditionally adapting an existing distance measure and thereby distorting the global data structure, random forest proximities provide a measure of local similarity which partially retains global information through the forest's recursive splitting process. Therefore, instead of exaggerating class separation, natural observational relationships are retained, as can be seen in low-dimensional visual representations (see Figure~\ref{fig:car-umap}). (2) Proximities formed using OOB data points retain the random forest's learning, thus, downstream task integrity is not jeopardized but relevant supervised information is retained. (3) Random forests are not limited to classification problems but also work with continuous labels. This provides an avenue for supervised metric learning in a regression context, as shown in Figure~\ref{fig:life_expectancy}. (4) A trained random forest model can extend similarity measures to unlabeled or out-of-sample observations, providing a means for semi-supervised metric learning or subsequent prediction. Additionally, noise variables are not likely to be used for splitting unless relevant variables are not included in the random subset of splitting variables. Subsequently, generated proximities naturally account for variable importance and can serve as a means of denoising.

We show that supervised manifold learning methods generally improve with the use of RF-GAP proximities. We compare the embedding mapping using common manifold learning algorithms including Isomap, $t$-SNE, Diffusion Map, Laplacian Eigenmaps, UMAP, PHATE, MDS, and Kernelized-PCA. Visualizations for each of these can be found in Appendix~\ref{sec:addtl_results}. Additionally, we demonstrate that diffusion-based embeddings generated using RF-GAP proximities better retain the random forest's learning in low dimensions, as shown in Figure~\ref{fig:knn_oob_diff}.

\section{Random Forest-Based Manifold Learning}\label{sec:rf-manifold}

Local connectivity is encoded via a distance metric. A kernel function, (e.g., Gaussian kernel) can be applied to the graph distances to provide a local measure of similarity between observations from which global relationships can be learned. For example, in diffusion processes, a stochastic matrix, or diffusion operator, is formed by row-normalizing the pair-wise similarities. The global structure is learned by powering the diffusion operator, simulating a random walk between observations. 

\begin{figure*}[!t]
    \centering
    \includegraphics[width = .8\textwidth]{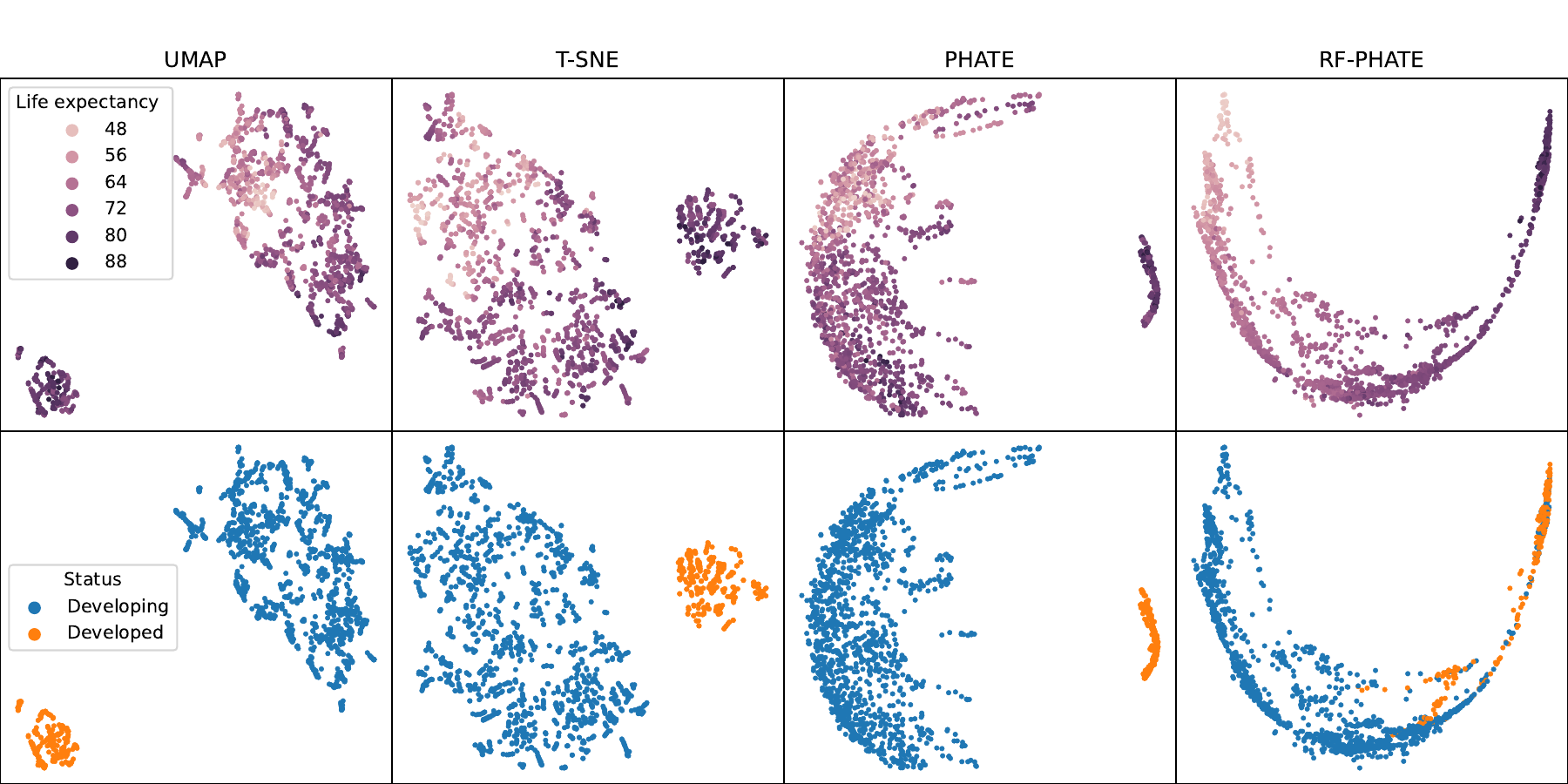}
    \caption{This figure demonstrates the ability of RF-GAP methods to work with continuous labels. The figures provided here are two-dimensional embeddings of the Life Expectancy dataset~\cite{naim2019life-expt} using UMAP, $t$-SNE, PHATE, and PHATE with the RF-GAP proximities, colored by life expectancy (top) and country economic status (bottom). The single most important variable for determining life expectancy is the country's economic status: developed or developing. All three unsupervised models completely separate these labels into two clusters, while RF-PHATE~\cite{rhodes2021rfphate} produces a progression incorporating these discrete labels in a continuous manner.}
    \label{fig:life_expectancy}
\end{figure*}

The quality of learned embedding is highly dependent on the kernel construction as well as global-structure mapping. Unlike an unsupervised kernel function, random forest proximities form noise-resilient, locally-adaptive neighborhoods ensuring that subsequent embeddings are constructed in a manner relevant to and consistent with the data labels. Similarities between points of different classes are still reflected in the proximity values, whereas this inter-class preservation is lost or diminished in class conditional measures such as the one given by Equation~\ref{eq:sisomap}. Continuous labels can also be reflected in the embedding using random forest proximities. We provide an example in Figure~\ref{fig:life_expectancy}. In this figure, embeddings are colored both by the life expectancy (the target label) as well as the country's economic status (developed vs. developing). It is clear that the unsupervised embeddings create separate clusters for each economic status, while the RF-PHATE embedding shows a continuum consistent with life expectancy in lower dimensions.

\section{Quantifying Results}\label{sec:results}

\begin{figure}[!htb]
    \centering
    \includegraphics[width = .4\textwidth]{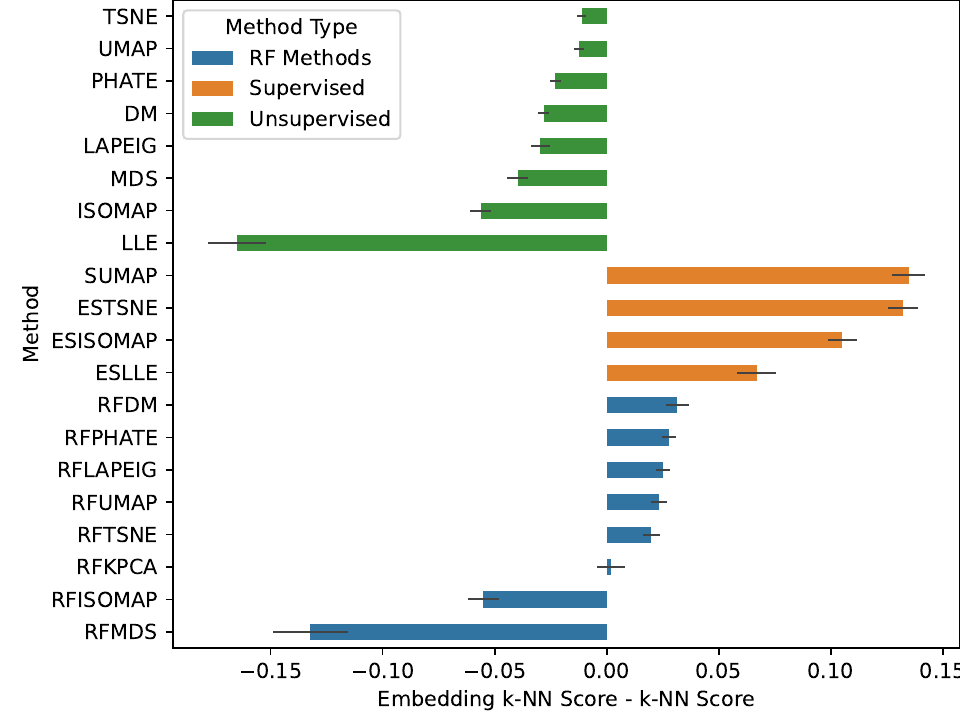}
    \caption{A $k$-NN model is trained on each embedding and the accuracies are compared with those of a model trained on the full dataset. Unsupervised embeddings tend to produce lower accuracies, while supervised embedding accuracies are inflated. Generally, RF-GAP-based embeddings produce accuracies more consistent with the full model. All $k$-NN accuracies were assessed using leave-one-out cross-validation. Score differences are aggregated across 20 random initializations across each dataset provided in Table~\ref{tab:datasets}}
    \label{fig:knn_diff}
\end{figure}

We use two methods to evaluate the low-dimensional embeddings in the supervised context. The first approach determines the extent to which the embedding can be used for the original classification problem. To this end, we use the 2-dimensional embeddings of unsupervised, supervised, and RF-GAP-based models to train a $k$-NN classifier to predict the original labels and compare the accuracy with that of a model trained on the full dataset. All accuracies were averaged using leave-one-out cross-validation and the overall results were aggregated across all datasets given in Table~\ref{tab:datasets}. Ideally, the difference in accuracies should be minimal, that is, we want to retain useful information without overfitting. In Figure~\ref{fig:knn_diff}, we see that the unsupervised embeddings produce accuracies lower than those using the full datasets, demonstrating a loss of information in low dimensions. The class-conditioned, supervised approaches tend to overfit the labels, generally producing much higher accuracies (near-perfect, in some examples) as a result of artificial class separation. The RF-GAP-based embeddings have accuracies more consistent with models trained on the full dataset, though results vary by method. The slightly higher values for the majority of RF-GAP methods can be accounted for by the superior predictive power of random forests over $k$-NN. We show in Figure~\ref{fig:knn_oob_diff} that the $k$-NN predictive accuracy of the RF-GAP embeddings typically aligns very well with the OOB score of the forest which generated the proximities, demonstrating the embeddings' abilities to retain the forest's learning in low dimensions.

\begin{figure}[!htb]
    \centering
    \includegraphics[width = .4\textwidth]{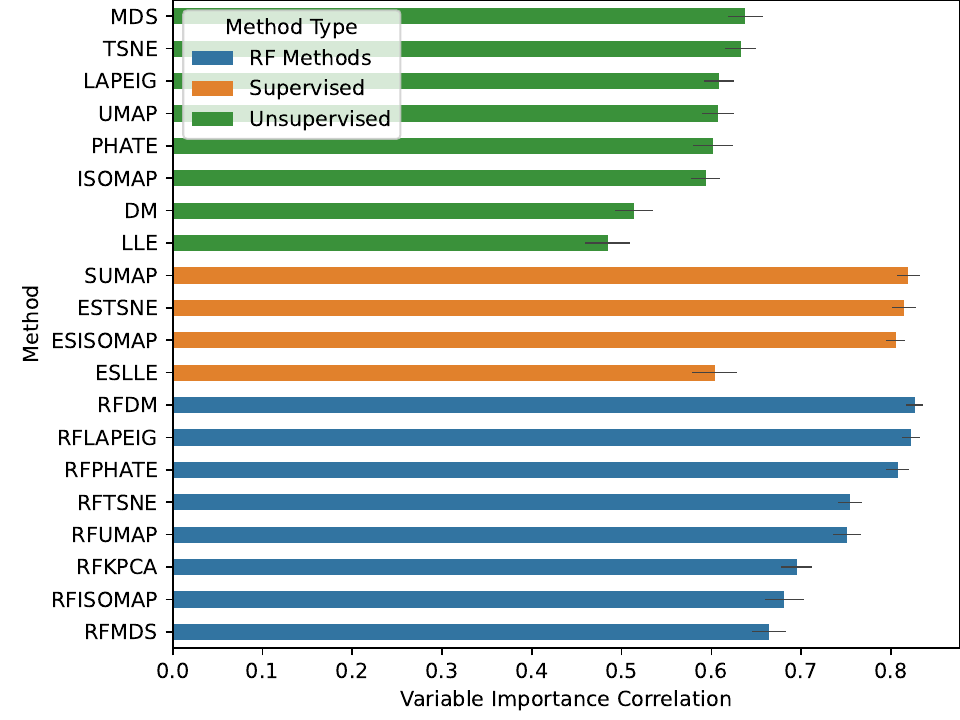}
    \caption{The correlation between importance scores shows to what extent variable importance is retained in the embedding. All RF-GAP methods outperform each of the unsupervised methods, and most other supervised methods have higher correlations than supervised methods. }
    \label{fig:imp_corr}
\end{figure}

The second evaluation technique provides an assessment of the hierarchy of variable importance retained in the embedding. In a supervised context, variables that provide higher class-discriminatory power are considered to be more important. For this evaluation, we first determine a permutation-based variable importance score using a $k$-NN classification model on the original supervised task. We then produce a second set of variable importance scores by regressing onto the embedding using a $k$-NN model trained on the original dataset. We calculate the correlation between the two variable importance scores. In Figure~\ref{fig:imp_corr}, we see that supervised models generally outperform unsupervised models in retaining variable importance, bearing in mind that class-conditional methods inflate class discrimination. Diffusion-based RF-GAP methods tend to perform best with this metric, suggesting they better preserve global, hierarchical variable importance.

\section{Conclusion}\label{sec:conclusion}

In this paper, we discussed the weaknesses of existing supervised manifold learning methods. We showed that RF-GAP-based manifold learning methods preserve local structure while maintaining global structure relative to data classes as can be seen in scatterplots of the embeddings. The visual quality of the RF-GAP embedding depends on the method used, but variable importance is maintained in low dimensions regardless of the method. Diffusion-based RF-GAP embeddings tend to retain the random forest's learning in low dimensions, suggesting that such methods better maintain the integrity of the kernel from which the embeddings were derived.




\bibliographystyle{IEEEtran}
\bibliography{IEEEabrv.bib, references.bib}

\clearpage
\appendices

\onecolumn
\counterwithin{figure}{section}
\counterwithin{table}{section}

\section{Additional Experimental Results}\label{sec:addtl_results}

\begin{figure}[!htb]
    \centering
    \includegraphics[width = .5\textwidth]{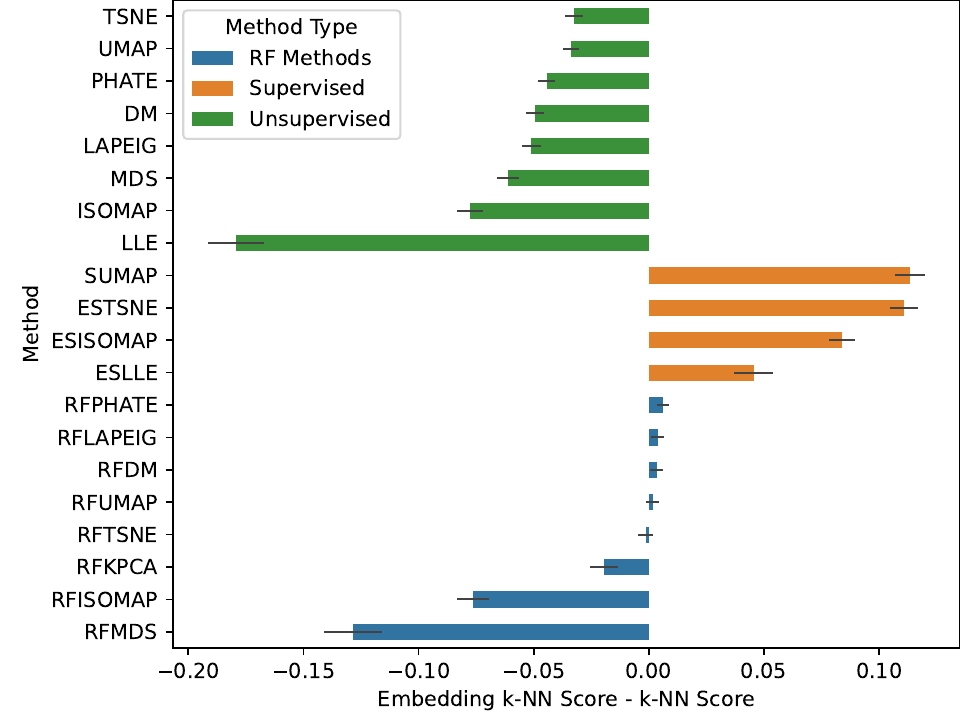}
    \caption{The difference between the OOB score from a trained random forest and the $k$-NN cross-validated score on the low-dimensional embedding. $k$-NN models trained on the RF-GAP diffusion-based embeddings as well as RFUMAP and RFTSNE embeddings nearly perfectly match OOB accuracies.}
    \label{fig:knn_oob_diff}
\end{figure}

The RF-GAP proximities perfectly preserve the random-forest predictive ability as was shown in~\cite{rhodes2022geometry}. In Figure~\ref{fig:knn_oob_diff}, we show that the predictive ability is well-preserved in low dimensions. In the figure, we see the diffusion-based methods as well as RFTSNE and RFUMAP methods nearly perfectly preserve the OOB score when using a $k$-NN classifier on the embeddings.  

The following figures provide examples to compare unsupervised, supervised, and RF-GAP-based methods. In most examples, the supervised models create near-linear separation between classes, while RF-GAP methods preserve relationships meaningful to the supervised task without global structure disruption.

\begin{figure}[!htb]
    \centering
    \includegraphics[width = \textwidth]{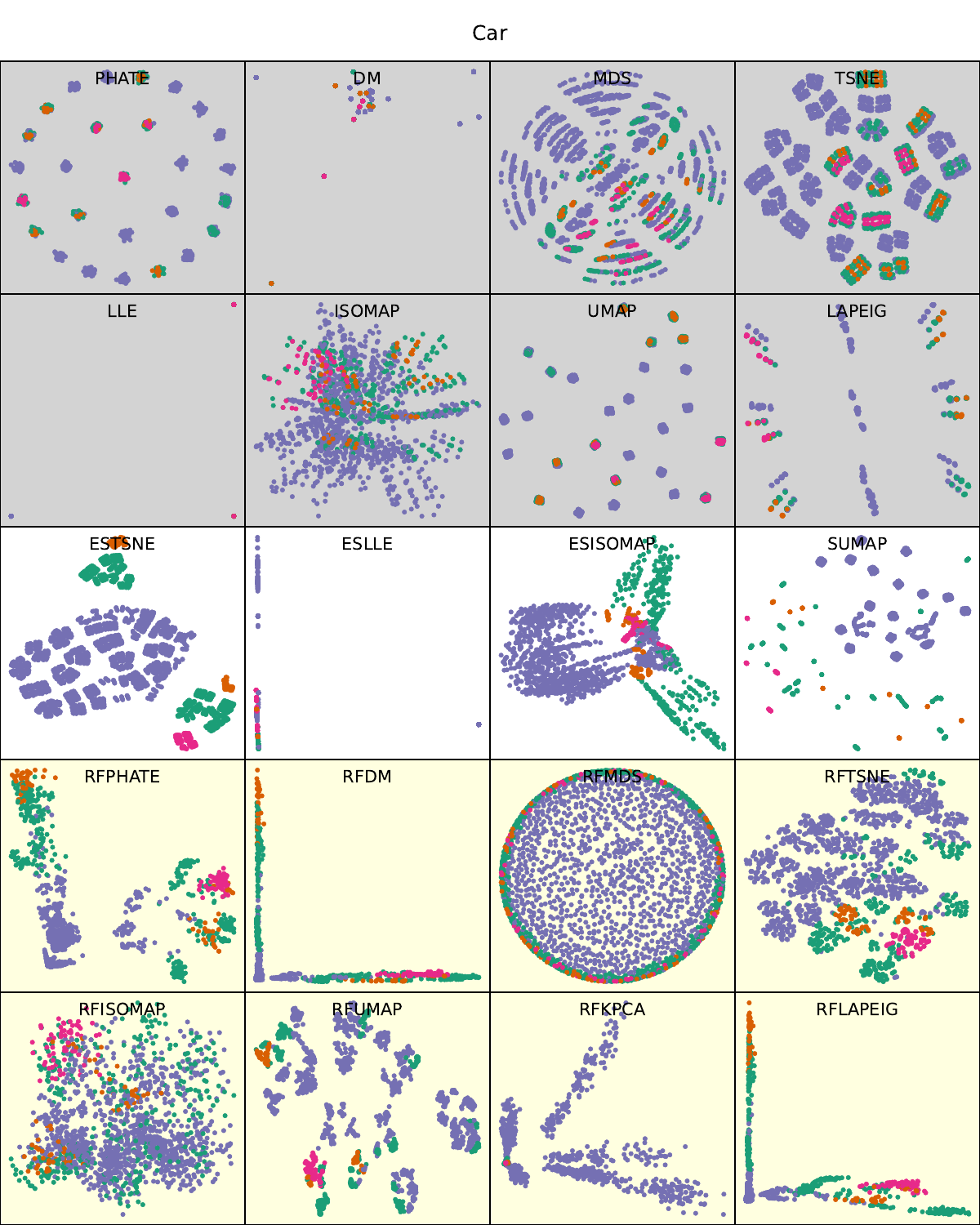}
    \caption{20 compared embeddings on the cars dataset~\cite{bohanec1988cars}. Embedding types are depicted by background color: grey for unsupervised, white for class-conditional supervised, and yellow for RF-GAP methods.}
    \label{fig:car-ALL}
\end{figure}

\begin{figure}[!htb]
    \centering
    \includegraphics[width = \textwidth]{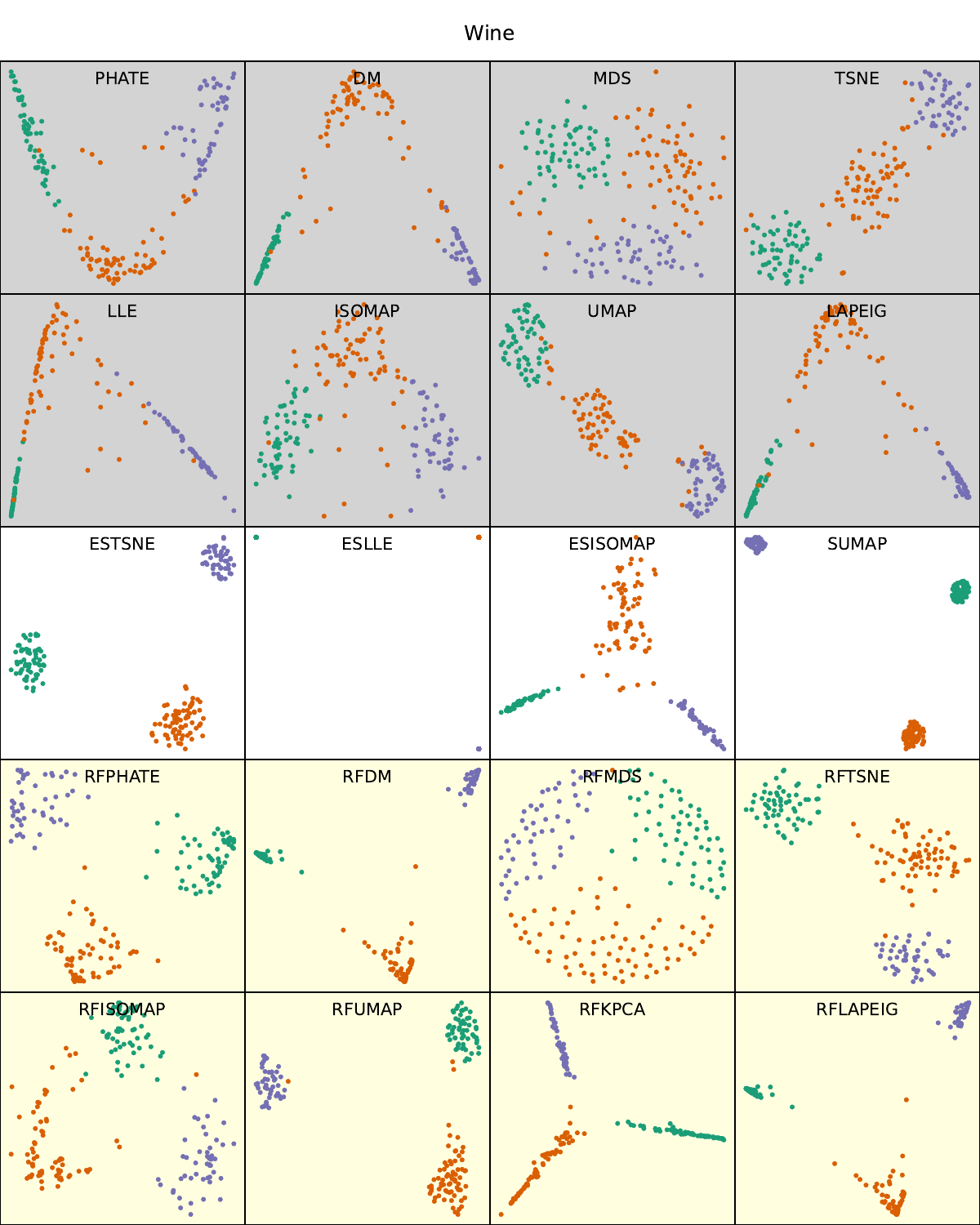}
    \caption{The wine dataset~\cite{UCI2019} is simple enough that using labeled information does not add much to the embedding.  At the same time, the RF-GAP methods are not diminished by the gained information from the random forest.}
    \label{fig:wine-ALL}
\end{figure}

\begin{figure}[!htb]
    \centering
    \includegraphics[width = \textwidth]{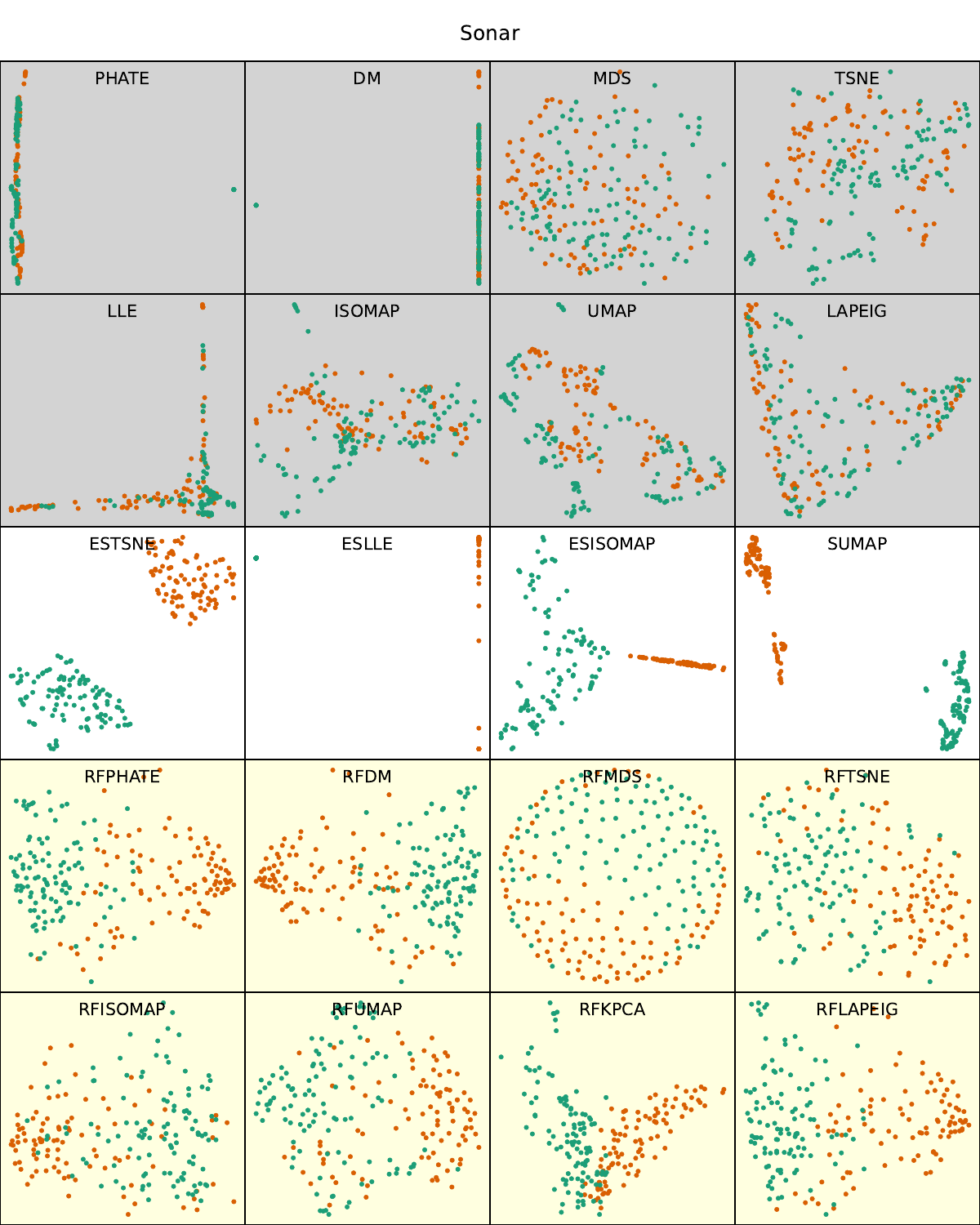}
    \caption{The sonar dataset~\cite{GORMAN198875sonar} reduced to two dimensions. The unsupervised models fail to find meaningful patterns, while the non-random forest supervised creates linear separation between classes. The RF-GAP models meaningfully retain the random forest model's learning.}
    \label{fig:sonar-ALL}
\end{figure}

\begin{figure}[!htb]
    \centering
    \includegraphics[width = \textwidth]{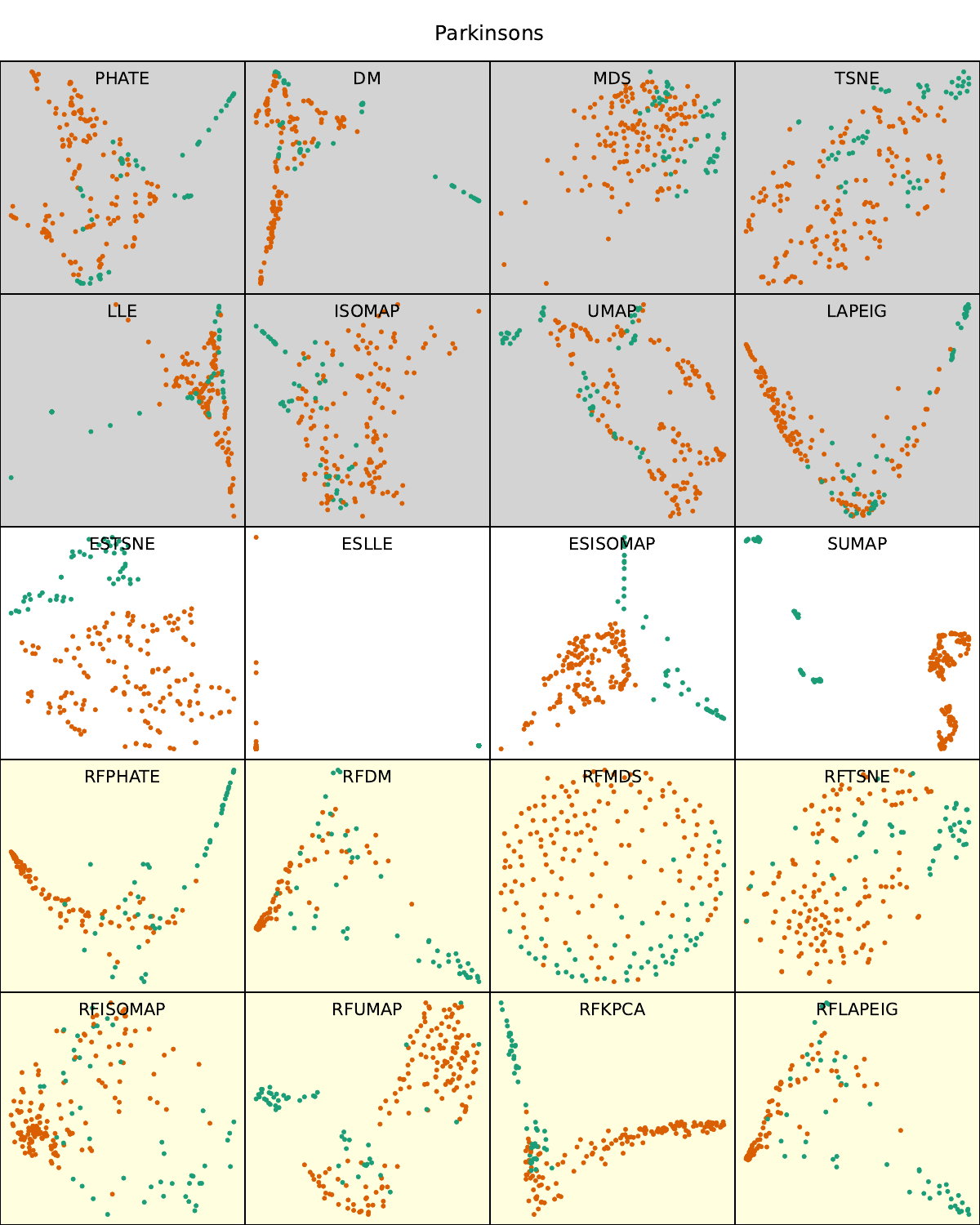}
    \caption{This example of the Parkinson's dataset~\cite{UCI2019} shows that the RF-GAP embeddings provide a somewhat clearer relationship between the two classes, while the other supervised methods provide perfect separation and the unsupervised methods contain extra noise.}
    \label{fig:parkinsons-ALL}
\end{figure}

\section{Datasets Used}\label{sec:datasets}

\begin{table}[!htb]
    \centering
    \caption{This table contains the descriptions datasets used in the quantitative experiments in Figures~\ref{fig:knn_diff}, \ref{fig:imp_corr}, and \ref{fig:knn_oob_diff}. Observations with missing values were removed, and uniquely-identifying variables were also removed. Each dataset was normalized before applying dimensionality reduction.}
    \begin{tabular}{|c|c|c|c|c|c|}
    \hline
        Data & Observations & Variables & Source  \\
    \hline
        Banknote  & 1372 & 5 & UCI \\
    \hline
        Breast Cancer   & 699 & 16 & UCI \\
    \hline
        Car  & 1728 & 6 & UCI \\
    \hline
        Diabetes  & 678 & 8 & UCI \\
    \hline
        E. Coli  & 336 & 8  & UCI  \\
    \hline
        Glass  & 214 & 10  & UCI  \\
    \hline
        Heart Disease  & 303 & 13 & UCI \\
    \hline
        Hill Valley  & 606 & 101 & UCI  \\
    \hline
        Ionosphere  & 351 & 34 & UCI  \\
    \hline
        Iris  & 150 & 4 & UCI  \\
    \hline
        Liver  & 345 & 7 & UCI \\
    \hline
        Lymphography & 148 & 18  & UCI  \\
    \hline
        Optical Digits & 3823 & 64 & UCI  \\
    \hline
        Parkinson's & 197 & 23 & UCI  \\
    \hline
        Seeds  & 210 & 7 & UCI  \\
    \hline
        Sonar & 208 & 60 & UCI  \\
    \hline
        Tic-Tac-Toe & 958 & 9 & UCI  \\
    \hline
        Waveform  & 5000 & 21 & UCI  \\
    \hline
        Wine  & 178 & 13 & UCI \\
    \hline
    \end{tabular}

    \label{tab:datasets}
\end{table}

\end{document}